\def\year{2017}\relax
\documentclass[letterpaper]{article}
\usepackage{aaai17}
\usepackage{amsfonts}
\usepackage{amsmath}
\usepackage{caption}
\usepackage{comment}
\usepackage{courier}
\usepackage{graphicx}
\usepackage{helvet}
\usepackage{lipsum}
\usepackage{multicol}
\usepackage{subfigure}
\usepackage{times}
\usepackage{url}

\begin{document}
\title{Active Discriminative Text Representation Learning}

\author{ {\bf Ye Zhang} \\
Department of Computer Science \\
University of Texas at Austin\\
yezhang@utexas.edu\\
\And
{\bf Matthew Lease}   \\
School of Information\\
University of Texas at Austin\\
ml@utexas.edu\\
\And
{\bf Byron C. Wallace}  \\
~~College of Computer \& Information Science\\
Northeastern University\\
byron@ccs.neu.edu\\
}

\maketitle
 
\begin{abstract}
\small
%\todo{Example todo note test!}
We propose a new \emph{active learning} (AL) method for text classification with convolutional neural networks (CNNs). In AL, one selects the instances to be manually labeled with the aim of maximizing model performance with minimal effort. Neural models capitalize on word embeddings as representations (features), tuning these to the task at hand. We argue that AL strategies for multi-layered neural models should focus on selecting instances that most affect the embedding space (i.e., induce discriminative word representations). This is in contrast to traditional AL approaches (e.g., entropy-based uncertainty sampling), which specify higher level objectives. 

We propose a simple approach for sentence classification that selects instances containing words whose embeddings are likely to be updated with the greatest magnitude, thereby rapidly learning discriminative, task-specific embeddings. We extend this approach to document classification by jointly considering: (1) the expected changes to the constituent word representations; and (2) the model's current overall uncertainty regarding the instance. The relative emphasis placed on these criteria is governed by a stochastic process that favors selecting instances likely to improve representations at the outset of learning, and then shifts toward general uncertainty sampling as AL progresses. Empirical results show that our method outperforms baseline AL approaches on both sentence and document classification tasks. We also show that, as expected, the method quickly learns discriminative word embeddings. To the best of our knowledge, this is the first work on AL addressing neural models for text classification.

%by introducing a stochastic interpolation between 

%We also propose a stochastic method for document classification that firstly considers both the embedding layer and the final representation (entropy), but later will focus more on final representation. Empirical results show that our method outperforms baseline AL approaches on three sentence datasets and three document datasets.
\end{abstract}

\section{Introduction}
\label{section:intro}

In \emph{active learning} (AL), the machine learning algorithm being trained is allowed to select the examples to be manually annotated by the teacher \cite{settles2010active}. The idea is that by selecting training data cleverly, rather than at i.i.d. random, better models can be learned with less effort, and thus at lower cost. This approach is attractive in scenarios in which labels are expensive but unlabeled data is plentiful.

%Active learning is a machine learning framework, where there are only few labeled data, and unlabeled data may be abundant. It allows the learner to select which instances to label, and then add to the labeled training set. A typical active learner starts with a small set of labeled instances, selects one or more informative query instances from a large unlabeled pool, learns from these labeled quires, and repeats this process. In this way, the learner aims to achieves high accuracy on the test set with as little labeling effort as possible. 

There has been a wealth of work on AL approaches for traditional machine learning methods in general~\cite{settles2010active}, and for text classification in particular \cite{tong2002support,mccallum98,wallace2010active}. However, almost no work has considered AL for text classification using modern neural models. We posit that the importance of \emph{representation learning} \cite{bengio2009learning} with neural models motivates exploring a rather different approach to AL for neural models vs.\ classic techniques.

In this work, we propose an AL method for convolutional neural networks (CNNs), which have recently achieved strong performance across many diverse text classification tasks \cite{kim2014convolutional,zhang2015sensitivity,johnson2014effective,zhang2016mgnc,zhang2016rationale}. 
%Ye_0730
These models first project words in texts into a low dimensional \emph{embedding layer}, and then apply convolution operations on the resultant matrix. 

While CNNs (and neural networks more generally) have demonstrated excellent performance when one has access to large amounts of training data, how can we make the best use of CNNs when annotation resources are scarce? Because word embedding estimation and tuning (for a specific text classification task) may be viewed as \emph{representation learning}, it is reasonable to optimize feature vectors before expending effort to tune the parameters of a model that accepts these as input. Indeed, adjusting the former will render updates to the latter potentially useless. Thus, we argue that the objective in AL (at least at the outset) should primarily be to select instances that result in better representations. %We note that this general AL approach would work with any model that estimates and exploits embeddings.

%CNN first projects the words in the text into a low dimension embedding layer~\cite{mikolov2013distributed}, and applies CNN on the embedding matrix representing the text~\cite{kim2014convolutional,zhang2015sensitivity,johnson2014effective,zhang2016mgnc}. This method has achieved very strong performance across many tasks. 

% bcw: again hoping that we can be a bit more general than sentiment *only* but can put back if that doesn't work out.
%It is natural to consider how to make the best use of CNNs when annotation resources are scarce; that is the question we address here. Specifically, we design a new AL strategy for CNNs.\footnote{We note that the general approach would work with any model that estimates and exploits embeddings.} We take the view that word embedding estimation and tuning (for a specific text classification task) is a form of \emph{feature learning}: thus the objective in AL should be to select instances that result in better features.%, i.e., embeddings.

More specifically, we propose a novel AL approach for sentence classification in which we select instances that contain words likely to most affect the embeddings. We achieve this by calculating the expected gradient length (EGL) \emph{with respect to the embeddings} for each word comprising the remaining unlabeled sentences. We show that this approach allows us to rapidly learn discriminative, task-specific embeddings. For example, when classifying the sentiment of sentences, we find that selecting examples in this way quickly pushes the embeddings of `bad' and `good' apart ({\bf Figure \ref{fig:embeddings-AL}}, bottom row). Ultimately, results show our AL method improves accuracy over several baseline AL approaches, across sentence and document classification tasks considered.

This method selects instances based on a max operator over the gradients expected for the individual words in a text, and thus is less appropriate for longer texts such as documents. Therefore, we extend our approach for document classification by linearly combining two scores: one corresponding to individual word embeddings and one measuring the overall uncertainty regarding instances.

%Ye_0730
\noindent In summary, key contributions of this paper include: 

\begin{itemize}

\item As far as we are aware, this is the first work to consider AL strategies explicitly for neural architectures in the context of text classification.

\item We demonstrate that variants of our model outperform baseline AL approaches that do not consider embedding-level parameters: on both sentence and document classification tasks our method realizes better performance with fewer labels, compared to baseline sampling approaches.

\item We also note that our approach substantially reduces the computational cost of AL, compared to previously proposed EGL approaches to AL. 

\end{itemize}
%This points to a potentially new line of work exploring active learning for deep/hierarchical models in natural language processing (NLP). 

\begin{comment}
Our approach is motivated by the following two observations: 
(i) In sentiment analysis tasks, only one or two words are important for the labels. For example, in a typical movie review 'This film is great', only 'is great' is important for the positive label of this sentence. So during the query phase, we prefer those unlabeled instances that have the most informative word. 
(ii) CNN has much more parameters to be learnt than the traditional machine learning method like support vector machine (SVM) or logistic regression. Though gradient-based method such as expected gradient length (EGL) achieves very strong results on traditional machine learning methods, using it as the query strategy in CNN makes it extremely expensive and impractical. Motivated by the above two points, we propose a new query strategy that can selects the instances only based on the informativeness of the word comprising the text, so that it not only reduces the computation to depend only on the first layer of CNN. Empirical results on sentiment tasks show that our model can achieve stronger learning curve than baseline methods, and can learn word representations better and more quickly. 
\end{comment}

\section{Preliminaries}

\label{section:preliminaries}

\subsection{CNNs for Text Classification}

We briefly review CNNs for text classification. Specifically we summarize the model proposed by Kim ~\shortcite{kim2014convolutional} and explored in depth by Zhang and Wallace \shortcite{zhang2015sensitivity}. We will denote the word \emph{embedding matrix} by $\mathbf{E}\in \mathbb{R}^{V\times d}$, where $V$ is the vocabulary size, and $d$ is the dimension of the embedding layer. A specific instance (sentence) is then represented by stacking the vectors corresponding to the words it contains (stored in $\mathbf{E}$), preserving word order. This results in an instance matrix $\mathbf{A}\in \mathbb{R}^{S\times d}$, where $S$ is the text length. 

\begin{figure}
\centering
\includegraphics[width=0.4\textwidth]{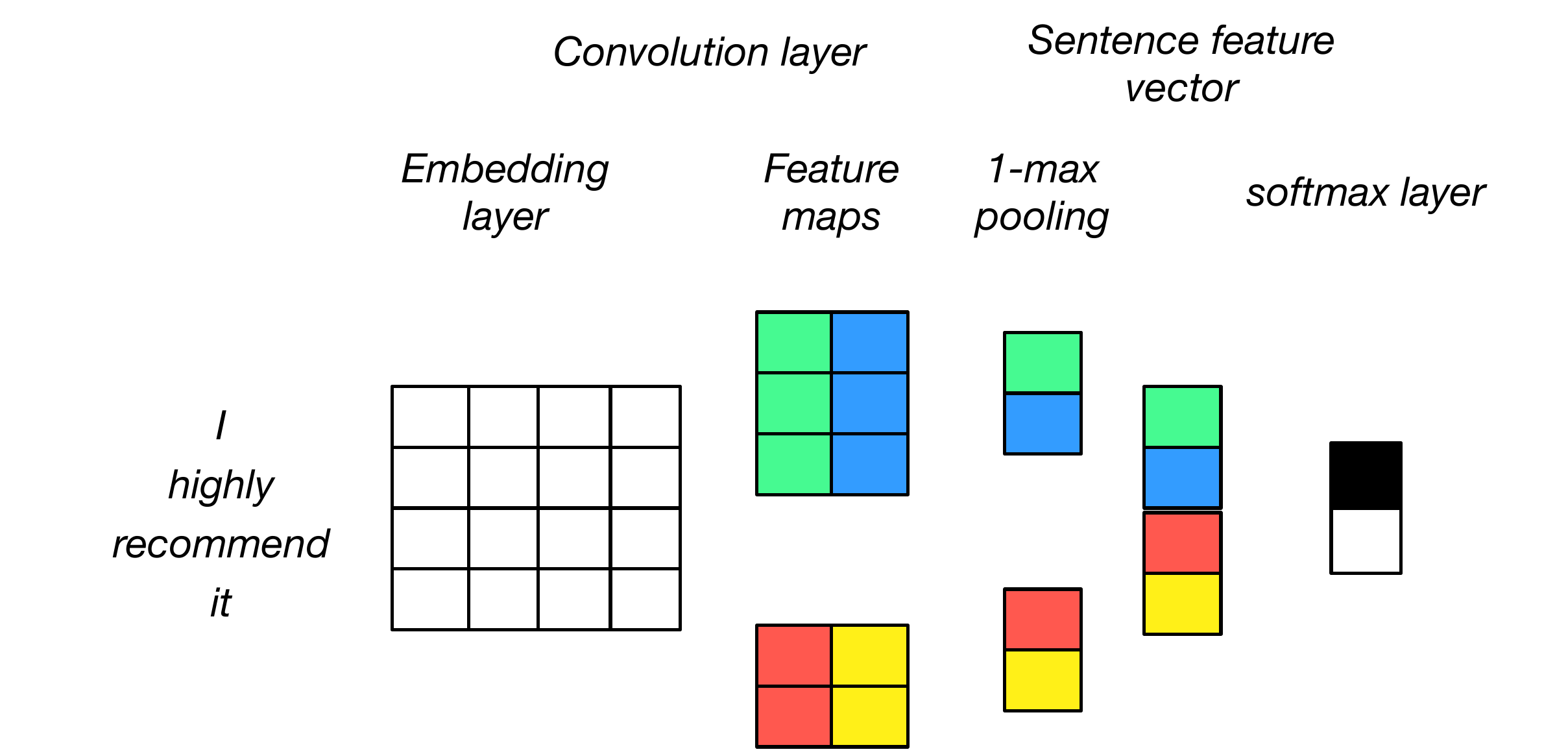}
%\vspace{-1em}
\caption{Illustrative schematic of a CNN for text classification. In this example, there are 2 feature maps with size 3, and 2 feature maps with size 2.} %Here there are four filters, two with heights 2 and two with heights 3, resulting in feature maps with lengths 3 and 2 respectively.}
\label{Single_CNN}
%\vspace{-1em}
\end{figure}

Convolution operations are then applied to this matrix, using multiple linear filters. Each filter matrix $\textbf{W}_i\in \mathbb{R}^{h\times d}$ performs a convolution operation on $\mathbf{A}$, generating a feature map $\mathbf{c}_i\in \mathbb{R}^{S-h+1}$. One-max pooling can then be applied to each $\mathbf{c}_i$ to obtain a feature value $o_i$ for this filter. (We note that we use multiple filter heights and redundant filters of each height.) Finally, all $o_i$ are concatenated to compose a final feature vector $\mathbf{o}$ for each instance. This is run through a softmax layer to induce a probability distribution over the output space. Typically this model is trained by minimizing the cross-entropy (or some other) loss via back-propagation~\cite{rumelhart1988learning}. {\bf Figure \ref{Single_CNN}} provides a schematic illustrating a toy realization of this model. For more details, see~\cite{kim2014convolutional,zhang2015sensitivity}. 

The above model for sentence classification can easily be generalized for document classification. In particular, we adopt the hierarchical approach described by Zhang et al. \shortcite{zhang2016rationale}, in which one first applies the above set of operations to each sentence comprising a document, and then sums these to induce a global representation, which is in turn fed through a softmax layer to obtain a final prediction.

\subsection{Active Learning}
\label{section:al}

%\subsection{Pool-Based AL}
We consider a \emph{pool-based} AL scenario~\cite{zhu2008active,tong2001support}, in which there exists a small set of labeled data $\mathcal{L}$ and a large pool of available unlabeled
data $\mathcal{U}$. The task for the learner is to draw examples to be labeled from $\mathcal{U}$ cleverly, so as to maximize classifier performance. These selections, or \emph{queries}, are typically made in a greedy fashion; an informativeness measure is used to score all candidate instances in the pool, and the instance maximizing this measure is selected.  %We describe specific strategies below.

%Queries are selectively drawn from the pool. Typically, instances are queried in a greedy fashion, according

%\subsection{Query Strategy Frameworks}

%All active learning scenarios involve evaluating the informativeness of unlabeled
%instances. In order to select queries, an active learner must have a way of assessing how informative each instance is. 

The key to developing AL strategies is designing a good informativeness measure. Let $\mathbf{x}^*$ be the most informative instance according to a \emph{query strategy} $\phi(\mathbf{x}_i;\boldsymbol\theta)$, or function used to evaluate each instance $\mathbf{x}_i$ in the unlabeled pool $\mathcal{U}$ conditioned on the current set of parameter estimates $\boldsymbol\theta$. We can define the following instance selection protocol:

\begin{equation}
    \mathbf{x}^* = \text{argmax}_{\mathbf{x}_i\in \mathcal{U}}\phi(\mathbf{x}_i;\boldsymbol\theta)
\end{equation}
%$\phi(x;\theta)$ is evaluated using the current model parameter $\theta$. 

For CNNs, $\boldsymbol\theta$ includes word embedding parameters $\mathbf{E}$, convolution layer parameters $\mathbf{C}$, and softmax layer parameters $\mathbf{W}$. 

Many querying strategies have been proposed in the literature \cite{settles2010active}. Our aim here is to ascertain whether AL works better in the case of neural models when one explicitly considers representation learning (i.e., focussing on $\mathbf{E}$); we selected the following three general baseline approaches because they enable us to explore this question directly.

%\subsubsection{Random Sampling}

%\vspace{.2em}

{\bf Random sampling}. This strategy is equivalent to standard (or `passive') learning; here the training data is simply an i.i.d. sample from $\mathcal{U}$. 

% bcw: removed for space -- "When using probabilistic models such as Naive Bayes or CNNs for binary classification, uncertainty sampling translates to selecting for labeling the instance with posterior probability nearest to 0.5. "
%\vspace{.2em}
%\subsubsection{Uncertainty Sampling}

{\bf Uncertainty sampling}. Perhaps the most commonly used query strategy is \emph{uncertainty
sampling}~\cite{lewis1994sequential,tong2002support,zhu2008active,ramirez:dmkd16}, in which the learner requests labels for instances about which it is least certain {\em wrt.} categorization. 

Uncertainty sampling can be instantiated in many ways, depending on the underlying classification model. A general uncertainty sampling variant uses entropy~\cite{shannon2001mathematical} as an uncertainty measure, defining $\phi(\mathbf{x}_i;\boldsymbol\theta)$ as: 
\begin{equation}
-\sum_k P(y_i=k|\mathbf{x}_i;\mathbf{\boldsymbol\theta})\text{log}P(y_i=k|\mathbf{x}_i;\boldsymbol\theta)
\label{equation:entropy}
\end{equation}

\noindent where $k$ indexes all possible labels.
Entropy-based uncertainty sampling often performs well \cite{settles2010active}. 

%\subsubsection{Expected Gradient Length}
% bcw: not sure that EGL is necessarily decision-theoretic? -- ah ok i sort of see; because we use the expectations... still wouldn't emphasize
% In this work, we use an “expected
%gradient length” (EGL) approach, which was introduced in~\cite{settles2008analysis} for active learning.
%\vspace{.2em}
\noindent {\bf Expected Gradient Length (EGL)}. This AL strategy aims to select instances expected to result in the greatest change to the current model parameter estimates when their labels are revealed (or provided) \cite{settles2008analysis}. The intuition is that one can view the magnitude of the resultant gradient as the value of purchasing a label; if this cost is small, then the label did not provide much new information. If the true class for a given instance were known, the gradient could be directly calculated under this assignment. But in practice this is unknown, and so the expectation is taken by marginalizing over the gradients calculated conditioned on possible class assignments, scaled by current model estimates of the posterior probabilities of said assignments. %Expected gradients are then calculated by marginalizing over these. 

% bcw: we need a name! maybe like expected embedding gradient length (EEGL)!?
\section{AL with CNNs/Embeddings}
\label{section:al-w-embeddings}

%%% bcw: maybe we should split this into the simple/sentence approach and then the doc-level approach, with corresponding subsections

We now introduce our proposed AL strategy for text classification with embeddings. This is based on the EGL method described above. In gradient-based optimization for neural models, the training gradient back-propagated to a set of model parameters given label $y_i$ for instance $\mathbf{x}_i$ may be viewed as a measure of change imparted by example $i$ for those parameters. Thus the learner should request the label for an instance expected to produce a large magnitude training gradient. If this gradient is taken with respect to all model parameters (distributed over all layers), then this is a straight-forward instantiation of EGL. Past work on EGL (involving linear models) adopted exactly this approach: the expected change to model parameters was evaluated over the entire set of parameters in $\boldsymbol\theta$. By contrast, we propose \emph{explicitly selecting examples that are likely to affect the representation-level parameters (i.e., the embeddings)}.

Formally, let $\nabla J(\mathcal{L};\boldsymbol\theta)$ be the gradient of the objective function $J$ with respect to the model parameters $\boldsymbol\theta$, where $J$ is the cost function. Further, let $\nabla J(\mathcal{L}\cup \langle \mathbf{x}_i,y_i\rangle;\boldsymbol\theta)$ be the new gradient that would be obtained by adding the training tuple $\langle \mathbf{x}_i, y_i \rangle$ to $\mathcal{L}$.
Because the true label $y_i$ will be unknown, we take an expectation over possible class assignments $k$. More precisely, we can calculate $\phi(\mathbf{x}_i;\boldsymbol\theta)$ as:
% bcw: should we consider us \mathbf for the x's? -- am putting in for now, i think it's more clear htis way
% bcw: i think we should use i's to index instances and then maybe k's for the actual class labels -- please double check what i've added :)
\begin{equation}
%\begin{align}
% bcw: please double check!
\sum_k P(y_i=k|\mathbf{x}_i;\boldsymbol\theta)\| \nabla J(\mathcal{L}\cup \langle \mathbf{x}_i,y_i=k \rangle;\boldsymbol\theta)\|
%\end{align}
\end{equation}

%But we propose instead taking the gradient w.r.t. the subset the word embedding parameters, thus explicitly maximizing the change in learned representations.

\noindent where $\|\mathbf{r}\|$ denotes the Euclidean norm of $\mathbf{r}$. Note that at query time $\|\nabla J(\mathcal{L};\boldsymbol\theta) \|$ should be near zero, assuming $J$ converged during the previous iteration. Thus, we can approximate $\nabla J(\mathcal{L}\cup \langle \mathbf{x}_i, y_i=k\rangle;\boldsymbol\theta) \approx \nabla  J(\langle \mathbf{x}_i,y_i=k\rangle;\boldsymbol\theta)$ for efficiency.
% bcw: i did not understand this comment? 
%, as training instances are assumed to be independent.
%Ye_19: Because they are independent, so we can decompose the total gradient into gradient of L and the gradient of the newly added single instance. And since gradient of L is zero, so we can need to only compute the gradient of the new instance. 

%  In traditional AL, all parameters $\boldsymbol\theta$ are assumed to be equally important. 
This approach selects instances that are likely to most perturb all model parameters $\boldsymbol\theta$. However, `deep' neural architectures are distinguished by their multi-layered structure, which corresponds to a large set of features distributed across different layers in the architecture. This makes calculating the EGL computationally expensive. More importantly, it is arguably incoherent to jointly consider the expected change \emph{at different layers} in the model. If we view lower levels in the model as learning to extract features, it makes little sense to jointly maximize expected change in these features \emph{and} to the parameters of the final softmax layer that accepts these as input. Changes to the former will immediately change the implications of perturbing the latter.

%We take the \emph{max} over the expected word gradients as the informativeness measurement of an instance because we are aiming to perturb particular word embeddings that are discriminative for the task at hand. 
Instead, we want to select unlabeled instances that can most improve the features learned by the model. Intuitively, it is paramount that the model learn good (discriminative) \emph{representations}; these will feed forward through the network, in turn improving classification. In the context of sentence classification --- in which instances comprise relatively few words --- we propose a querying strategy that scores sentences using the maximum expected gradient over the words they contain. In the case of longer texts or documents (which contain many words), it is intuitive to strike a balance between myopically selecting instances to maximize individual word gradients on the one hand, and considering the model's overall uncertainty regarding the instance on the other. We next elaborate on the methods we propose for these two scenarios.

%======================================================
\subsection{Active Sentence Classification with CNNs}

{\bf {\em EGL-word} model.} For sentence classification, we adopt the following as our scoring function for sentence classification. For each instance (sentence) in $\mathcal{U}$, we take the expected gradient with respect to \emph{only} the embeddings of its constituent words, selecting the example that maximizes this expected embedding gradient as our measure of informativeness. Intuitively, we use a max-over-words approach to adjust particular word embeddings that are discriminative for the task at hand. Formally, we define our $\phi(\mathbf{x}_i;\boldsymbol\theta)$ as:
%\vspace{-1em}
% bcw: please double check; i've amended so that j runs over x_i ... which i think is correct
\begin{equation}
 %\text{max}_{j\in \mathbf{x}_i}
 \max_{j\in \mathbf{x}_i} \sum_{k}P(y_i=k|\mathbf{x}_i;\boldsymbol\theta)\| \nabla J_{\mathbf{E}^{(j)}}(\langle \mathbf{x}_i,y_i=k \rangle; \boldsymbol\theta)\|
 \label{eq:EGL-word}
\end{equation}

\noindent Where we denote by $\nabla J_{\mathbf{E}^{(j)}}$ the gradient of $J$ with respect to the embedding of word $j$ ($j$ ranges over the words in $\mathbf{x}_i$). Note that the gradient is only taken for each word in the instance $\mathbf{x}_i$. The gradients for embeddings corresponding to words not in $\mathbf{x}_i$ are 0 and can thus be ignored; this is a computational boon because instances tend to be sparse. %We refer to this strategy as \textit{EGL-word}, and 
%We adopt \textit{EGL-word} directly as our scoring function for sentence classification tasks. 
Another straightforward strategy to measure the informative of a sentence is to replace the `max' operator in equation \ref{eq:EGL-word} with the average operation. That is, instead of choosing the word with the maximum expected gradient, we can average on the expected gradients of all the words in the sentence. But this method does not work as well as \textit{EGL-word}. We attribute this to the fact that in a short sentence, most words are not relevant to the label of the sentence. %Ye_1117

%\vspace{.2em} 

\noindent{\bf {\em EGL-sm} model.} Whereas {\em EGL-word} focuses on parameters associated with the lowest level in the model ({\bf Figure \ref{Single_CNN}}), we also consider the other extreme in sentence classification tasks: taking the gradient with respect to only the final softmax layer parameters $\mathbf{W}$. In this case $\phi(\mathbf{x}_i;\boldsymbol\theta)$ becomes: 
\begin{equation}
\sum_k P(y_i=k|\mathbf{x}_i;\boldsymbol\theta)\|\nabla J_{\mathbf{W}} (\langle \mathbf{x}_i,y_i=k \rangle ;\boldsymbol\theta) \|
\end{equation}
\noindent where $J_{\mathbf{W}}$ denotes the gradient {\em wrt}.\ %with respect to 
the softmax layer. %We refer to this strategy as \textit{EGL-sm}. 

\subsection{Active Document Classification with CNNs}

\noindent {\bf {\em EGL-word-doc} model.} For longer text classification tasks, we modify the above \textit{EGL-word} variant in a few key ways. First, we normalize the gradient of each word by dividing it by its frequency in the document. This is because in longer texts there exist many `stop words' such as `the', and their gradients dominate if occurrence counts are ignored, since there are more branches flowing back to these words during back-propagation. %Ye_1117 
Accounting for term frequencies in the gradient calculation mitigates this issue. Second, rather than exclusively relying on the single word with the largest gradient to score documents, we sum over the (frequency-normalized) gradients corresponding to the top $k$ words. The number of top words ($k$) is a hyper-parameter and will depend on the average document length in a given corpus. 
%\todo{Should we still call this EGL-word or is it different enough to have a different name?  If not that different, then don't need this "we refer to" sentence.} 
%Ye_11: Matt -- This method is a variant from 'EGL-word in sentence tasks', and the basic idea is the same. We use "We refer to this method as EGL-word for document classification" as to distinguish it from EGL-word in sentence classification, and let the reviewer know that we use the same name as in sentence tasks, though with some variants. But if it is confusing, maybe rename it to EGL-word-norm? (since we do normalization).
% bcw_13: perhaps we should renam eto EGL-word-norm to be explicit
%Ye_13: I'll use EGL-word-doc to distinguish it from EGL-word in sentence. 
We refer to this method as \textit{EGL-word-doc} for document classification.\footnote{Experiments  applying the same variant of \textit{EGL-word}  used for sentence classification does not perform as well for longer texts. \textit{EGL-sm} model also performs much worse than the other methods in the document classification tasks, so we do not report their results. }%Ye_1117 
%Ye_12: Note that EGL-word reported the for document task in the figure and table is 1) normalized by term frequencies 2) considering top k words rather than only top words, so it's different from EGL-word in sentence task. I did not report the same EGL-word method for sentence task on document task. 

%% bcw: ye -- why will the gradients of stop words dominate though? wouldn't we expect these updates to be small? or i guess we're summing over the number of times it occurs or something?
%Ye_09: But stop words like 'and' and '.' appear multiple times in the document, so I think the gradient will be larger for these words if we don't do normalization. 

%In document tasks, we use random sampling an entropy based sampling as baselines. 

{\bf {\em EGL-Entropy-Beta} model.} In addition to the above modifications, we extend our approach for longer text classification to jointly consider: (1) the expected updates to word gradients (for words in the instance); and (2) the current uncertainty regarding the instance. For the former, we use \textit{EGL-word-doc} (modified as described above), and for the latter we use entropy (Equation \ref{equation:entropy}). We denote the entropy score by $\phi_{\text{Entropy}}$ and the \textit{EGL-word-doc} score by $\phi_{\text{EGL-word-doc}}$. We interpolate these to form a composite document score. 

These scores are on incomparable scales, so we normalize them by transforming them into percentiles. $\mathcal{P}$($i$, $\mathcal{U}$) is used to denote the percentile of the score of a given instance among a pool of instances $\mathcal{U}$. For example, $\mathcal{P}$($i$, $\mathcal{U}$)=87\% indicates that 87\% of the instances in $\mathcal{U}$ are smaller than $i$. To encode the relative entropy score of a given instance in $\mathcal{U}$, we use $\mathcal{P}$($\phi_{\text{Entropy}}(i), \{\phi_{\text{Entropy}}(j):j\in{\mathcal{U}}\}$). We can now define our composite, interpolated scoring function which considers feature learning and output certainty jointly:

%Then we propose our new score function for instance $i$ incorporating both entropy and EGL-word in iteration $t$ as follows:
%\vspace{-10pt}
\begin{equation*}
\begin{split}
\phi_t(i) = \gamma_t \cdot \mathcal{P}(\phi_{\text{Entropy}}(i), \{\phi_{\text{Entropy}}(j):j\in{\mathcal{U}}\}) +\\
(1-\gamma_t) \cdot \mathcal{P}(\phi_{\text{EGL-word-doc}}(i), \{\phi_{\text{EGL-word-doc}}(j):j\in{\mathcal{U}}\})
\end{split}
\end{equation*}

We treat the interpolation parameter $\gamma_t$ --- constrained to be between 0 and 1 --- as a random variable with a temporal dependence ($t$ indexes time, or AL iteration). Intuitively, we assume that at the outset of AL, the model should pay relatively more attention to learning discriminative representations of words. As learning progresses, focus should shift toward the higher-level uncertainty-based score. To realize this intuition, we assume $\gamma_t\sim \text{Beta}(\alpha,\beta_t)$. We decrease $\beta_t$ linearly over time (AL iterations), which has the desired effect of increasing the expectation of $\gamma_t$, in turn increasing the attention paid to the document level entropy score. 
We found that drawing $\gamma_t$ from a distribution yields smoother performance compared to setting it deterministically. %Ye_1117

\begin{table}%[h!]
\centering
%\footnotesize
\begin{tabular}{c | c c c} 
%\hline  \\[-.8em]
  & CR & MR & Subj \\ 
 \hline \\[-.5em]
 Positive & 2406  & 5331  & 5000 \\ 
Negative & 1367 & 5331 & 5000 \\
 Avg. word count & 19 & 20 & 23 \\
 %\hline
\end{tabular}
\caption{Statistics of sentence datasets.}
\label{sentence_data}
%\vspace{-1em}
\end{table}

\begin{table}%[h!]

\centering
%\footnotesize
\begin{tabular}{c | c c c} 
 %\hline  \\[-.8em]
  & MR & MuR & DR \\ [0.25ex] 
 \hline  \\[-.5em]
 Positive & 1000  & 1000 & 23649  \\ 
 Negative & 1000 & 1000 & 30254 \\
 $l_{\text{sen}}$ & 21.2 & 16.8 & 15.2 \\
 $l_{\text{doc}}$ & 32.6 &7.5 & 4.1\\
 %\hline
\end{tabular}
\caption{Statistics of document datasets. $l_{\text{doc}}$ denotes the average sentence length in words, and $l_{\text{doc}}$ denotes the average number of sentences per document.}
%\vspace{-10pt}
\label{doc_data}
\end{table}

%%%%%%%%%%%%%%%%%%%%%%%%%%%%%%%%%%%%%%%%%%%%%%%%
%%%%%%%%%%%%%%%%%%%%%%%%%%%%%%%%%%%%%%%%%%%%%%%%
\section{Experimental Setup}
\label{section:experiments-and-results}
% bcw: We have more space now -- so as per Matt's comment, let's give more details on these?
% bcw_13: just kidding we don't have space.

%\subsection{Datasets}

We report results on three sentence datasets and three document datasets. {\bf Tables \ref{sentence_data}} and {\bf \ref{doc_data}} provide key statistics for each dataset. We briefly describe each dataset below and refer the reader to the source citations for additional details.

%%%
% bcw: @TODO -- extend description; at least summarize the task. Possibly also include size, average sentence/doc length and class breakdowns? I think would be helpful. Possibly we could even do two tables? 
%Ye_0908: Done

~\\
\noindent {\bf Sentence Datasets}\\
\noindent \textbf{CR}: {\em positive}\,/\,{\em negative} product reviews   \cite{hu2004mining}.\footnote{%\scriptsize 
\url{www.cs.uic.edu/∼liub/FBS/sentiment-analysis.html}}
 
\noindent \textbf{MR}: {\em positive}\,/\,{\em negative} movie reviews \cite{pang2005seeing}. %\footnote{https://www.cs.cornell.edu/people/pabo/movie-review-data/}

\noindent \textbf{Subj}: {\em subjective}\,/\,{\em objective} sentences  \cite{pang2004sentimental}.\footnote{MR and Subj datasets are available at: {\url{ http://www.cs.cornell.edu/people/pabo/movie-review-data/}}. }
%\end{itemize}

~\\
\noindent {\bf Document Datasets} {\em positive} / {\em negative} classification tasks \\
%\subsubsection{Document Datasets}
%\hfill \break
\noindent \textbf{MR}: (Longer) movie reviews
\cite{pang2004sentimental}\footnote{Both MR datasets can be found online at the same URL.}.
%\scriptsize \url{www.cs.cornell.edu/people/pabo/movie-review-data/}}
%Ye_11: "same as sentence dataset" is quite confusing. The reviewer might think they are the same dataset. 
%bcw_13: changed -- agree that's very confusing

\noindent \textbf{MuR}: Music reviews  %labeled as `positive' or `negative' 
\cite{blitzer2007biographies}.\footnote{\url{http://www.cs.jhu.edu/~mdredze/datasets/sentiment/}}

\noindent \textbf{DR}: Doctor reviews \cite{wallace2014jamia}.%\todo{public data?}
% bcw_13: it is, yes, but the link in orig paper (to Brown) is dead and people have to email me for it now... 
  %labeled as `positive' or `negative' 
%Ye_09: Byron, is there any reference for this dataset? 
% bcw: yes! added
%Ye_11: if space is not enough, maybe remove dataset statistics 
%\begin{comment}
%The statistics of these document datasets are shown in Table \ref{doc_data}.

\begin{figure}%[ht!]
\centering
\includegraphics[width=0.425\textwidth]{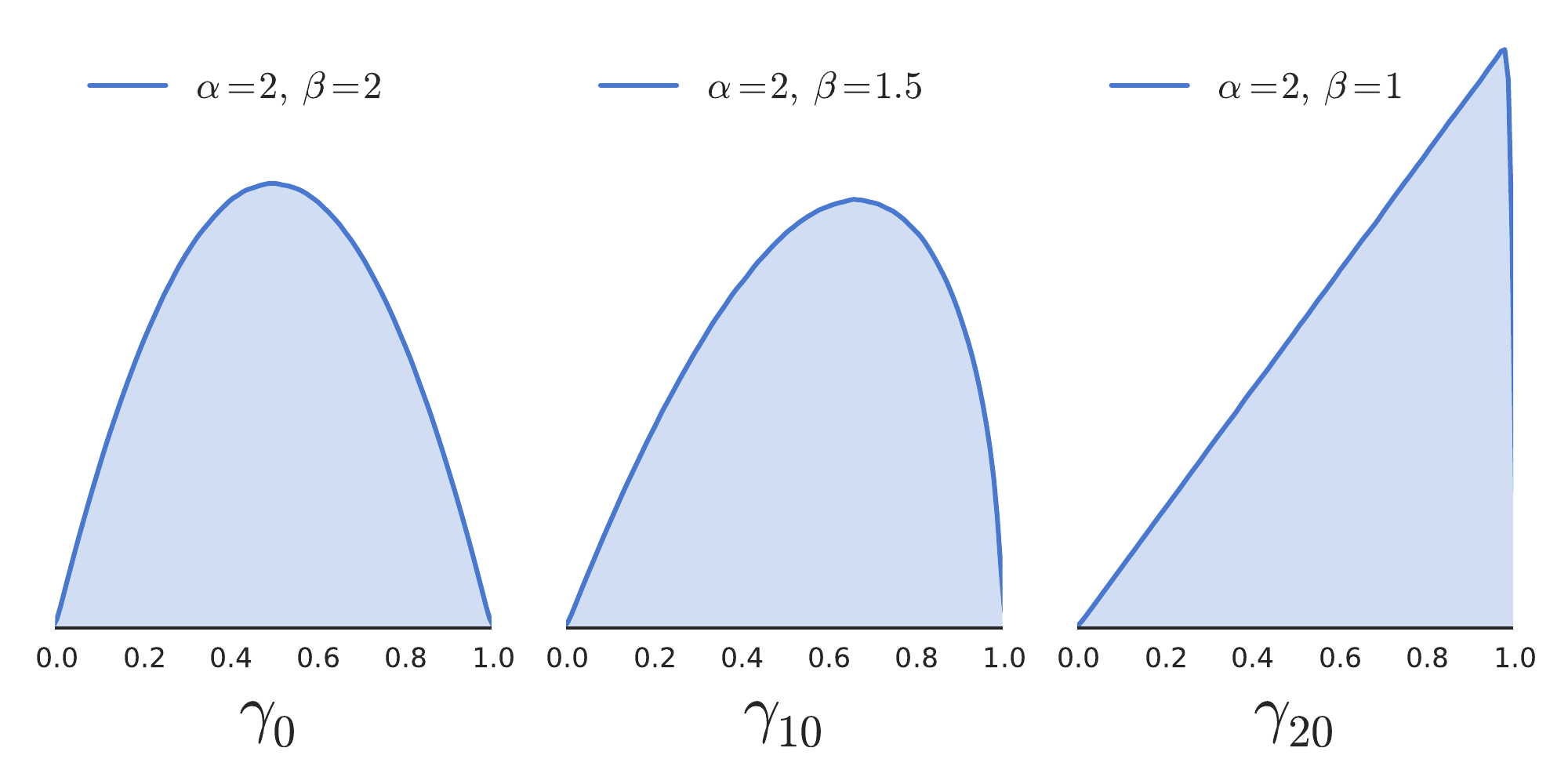}
%\vspace{-1em}
\caption{Beta distributions over $\gamma_t$ at $t$=0, $t$=10, $t$=20.}
\label{beta}
%\vspace{-1em}
\end{figure}

%-------------------------------------------------
\subsection{Model Configuration}
\label{section:model-config}

% bcw: ye, please double check my rationale below `we used more for document classification tasks because we expect more diversity' -- is that about right? if not, why did we use 50 for sentence and 100 for doc? 
%Ye_0908: I think it's just because I directly modify from the doc-CNN code, and I forget to change it to 50... . But I don't think 50 and 100 will make big difference. And your explanation is reasonable. 
We used standard pre-trained {\tt word2vec}-induced vectors\footnote{\url{https://code.google.com/archive/p/word2vec/}} to initialize $\mathbf{E}$. As per Zhang and Wallace \shortcite{zhang2015sensitivity}, we used three filter heights (3, 4, 5). For sentence and document classification tasks, we used 50 and 100 filters of each size, respectively.\footnote{We used more filters for document classification tasks because we expect more diversity in longer pieces of text, but we found that the performance was not sensitive to this choice in any case.} Given our goal to explore AL strategies appropriate for neural architectures (particularly CNNs), rather than to maximize absolute CNN performance for new state-of-art results, we did not tune these hyperparameters.   %compared to previous methods on particular datasets, 
%but rather to . %And to make the comparison with traditional AL methods fair, we keep their hyperparameter the same. 

% bcw: possibly show plots similar to what I sent along showing the distribution of the Beta distribution with various params... assuming there's room... 
We performed 20 rounds of batch active learning. %Ye_19: 
At the outset, we provided all learners with the same 25 instances (sampled i.i.d. at random). In subsequent rounds, each learner was allowed to select 25 instances from $\mathcal{U}$ according to their respective querying strategies. These examples were added to $\mathcal{L}$, and the models were retrained. % on this augmented training set. 

For \textit{EGL-word-doc} and \textit{EGL-Entropy-Beta} in document classification, the number of top words $k$ used to calculate the score for each document was set to 3, 2 and 30 respectively for MuR, DR, and MR datasets. 
For \textit{EGL-Entropy-Beta}, we fixed $\alpha=2$ and  initialized $\beta=2$ as well, which implies a roughly equal weight on embedding and uncertainty scores. We then decreased $\beta_t$ linearly with iterations $t$. Thus $\gamma_t$ is expected to increase over time, ascribing more weight to the entropy score. 
%\begin{comment}
For reference, {\bf Figure \ref{beta}} provides illustrative empirical distributions used for $\gamma$ at three time points during AL. To reiterate, our goal was to shift from initially paying equal attention to the representation learning and instance uncertainty criteria, to increasingly focusing on the latter (document-level uncertainty) as time progresses. 

%Ye_12: But the figure for beta distribution does not mean "shift emphasis from feature learning". At the beginning, it puts equal weight on feature learning and document-level uncertainty. 
% bcw_13: Sure; at the beginning the focus is equal, but this changes to being biased toward uncertainty sampling from this baseline - have reworded to clarify

%We give some sample distribution over time
%in Figure \ref{beta}.

We evaluated performance by calculating accuracy (classes are fairly balanced)
%\footnote{
%Ye_11: explain why only report accuracy 
%Accuracy is an appropriate metric here because classes are reasonably balanced.} 
%bcw_13: please double check what I wrote here is accurate, Ye
on a held-out test set after each round. For all but one dataset we repeated this entire AL process 10 times, using test sets generated via 10-fold CV. The exception was the doctor reviews (`DR') dataset, which is comparatively large; we therefore used a single big test set in this case. We replicated all experiments 5 times for all train/test splits, for all datasets, to account for variance. 
%Ye_13: Yes, correct!
%We datasets save for the doctor reviews (`DR') document classification dataset, which has a comparatively big test set. In this case, we used only  replicating the experiment 5 times per fold to account for variation. Reported results are means of these averages. 
% bcw_13: so for DR did we just do it once? 
We estimated parameters by Adadelta \cite{zeiler2012adadelta}, tuning $\mathbf{E}$ in back-propagation to induce discriminative embeddings.
%Ye_24: repeated the entire AL process 10 times? It's 10-fold CV, and for each fold, 5 replications. 

%For this we used batch sizes of 25, and ran for 10 epochs. Performance was evaluated by calculating accuracy on a held-out test set after each iteration of AL. We repeated the entire AL process 10 times (10-fold CV), and for each fold we replicated the experiment 5 times to account for variation; reported results are averages of these. 

%The accuracy is by taking the average across 10-fold CV, and for each CV, we replicate the experiments 5 times, and again take the average of the results. 

\section{Results and Discussion}
\label{section:results}

\begin{figure*}%[!t]
%\vspace{-2em}
\begin{multicols}{3}
    \includegraphics[width=\linewidth]{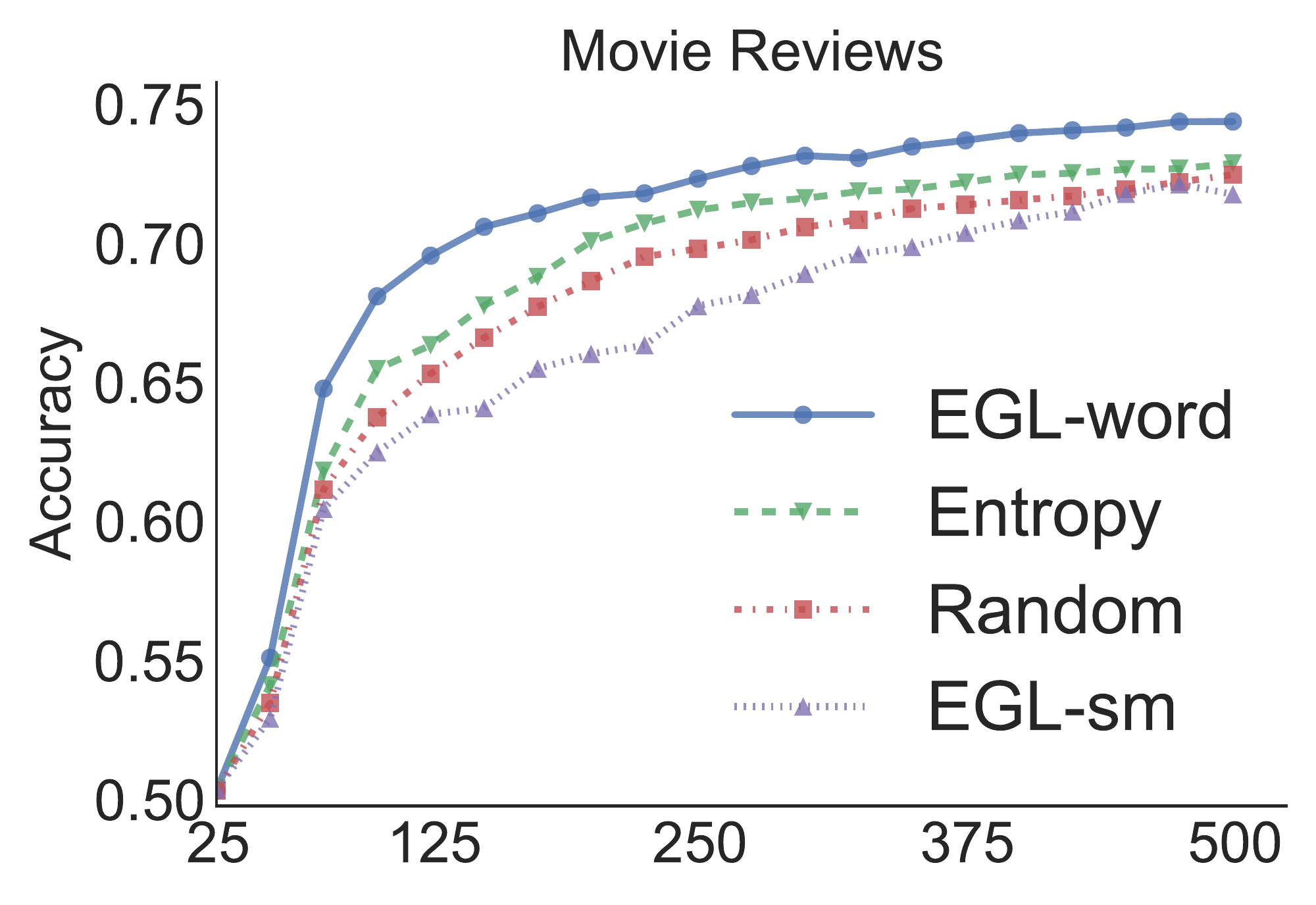}
    \includegraphics[width=\linewidth]{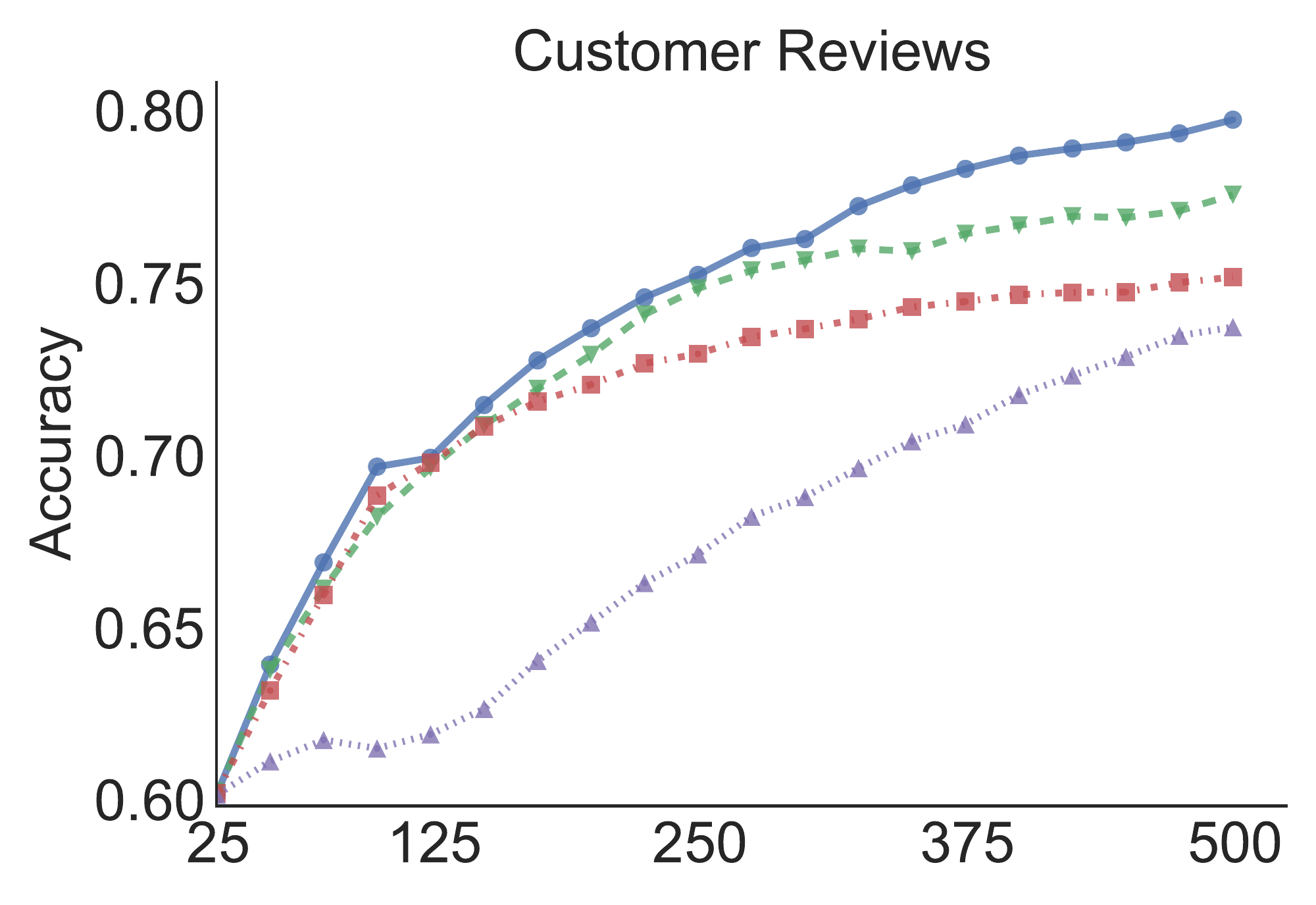}
    \includegraphics[width=\linewidth]{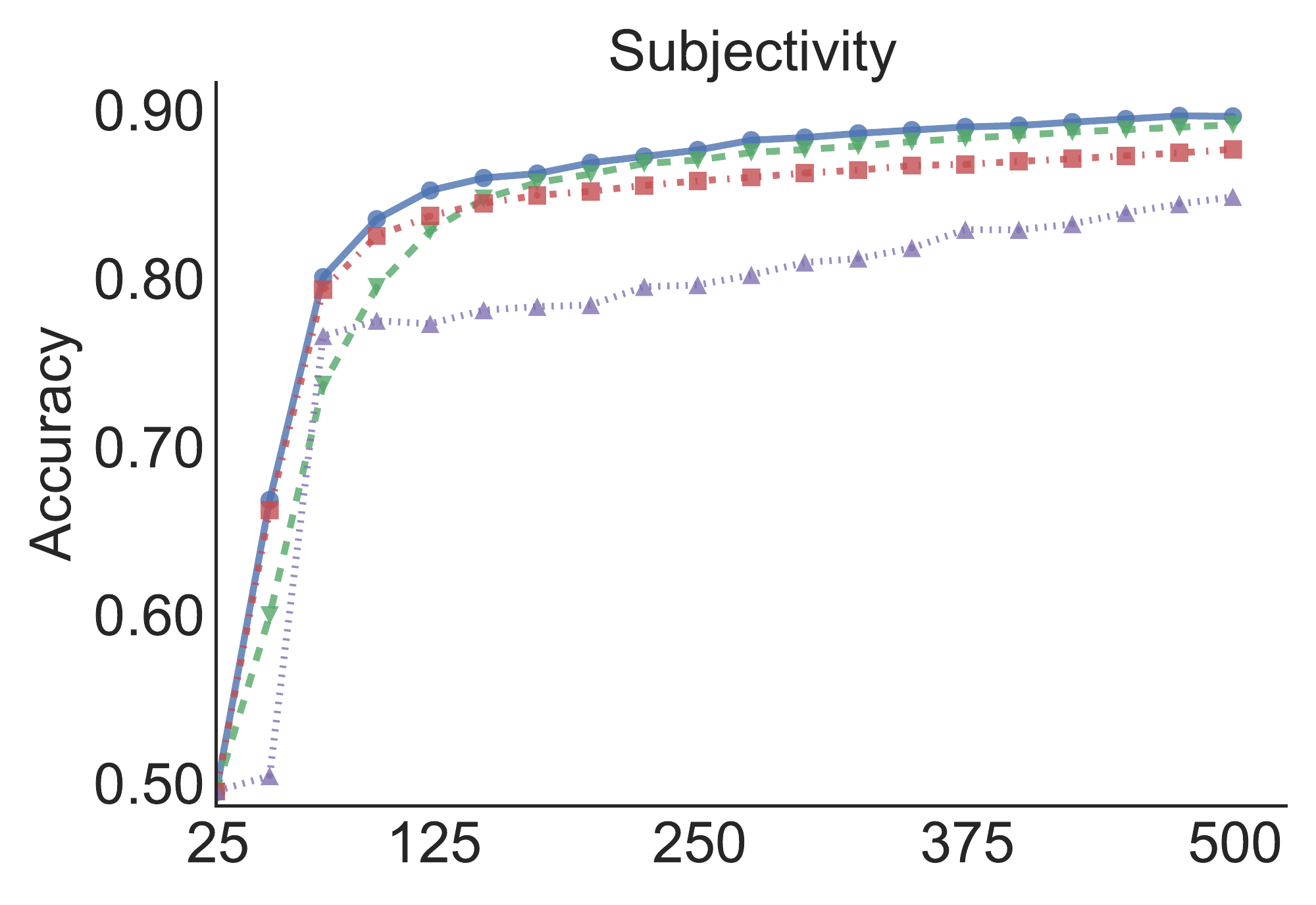}
\end{multicols}
\begin{multicols}{3}
    \includegraphics[width=\linewidth]{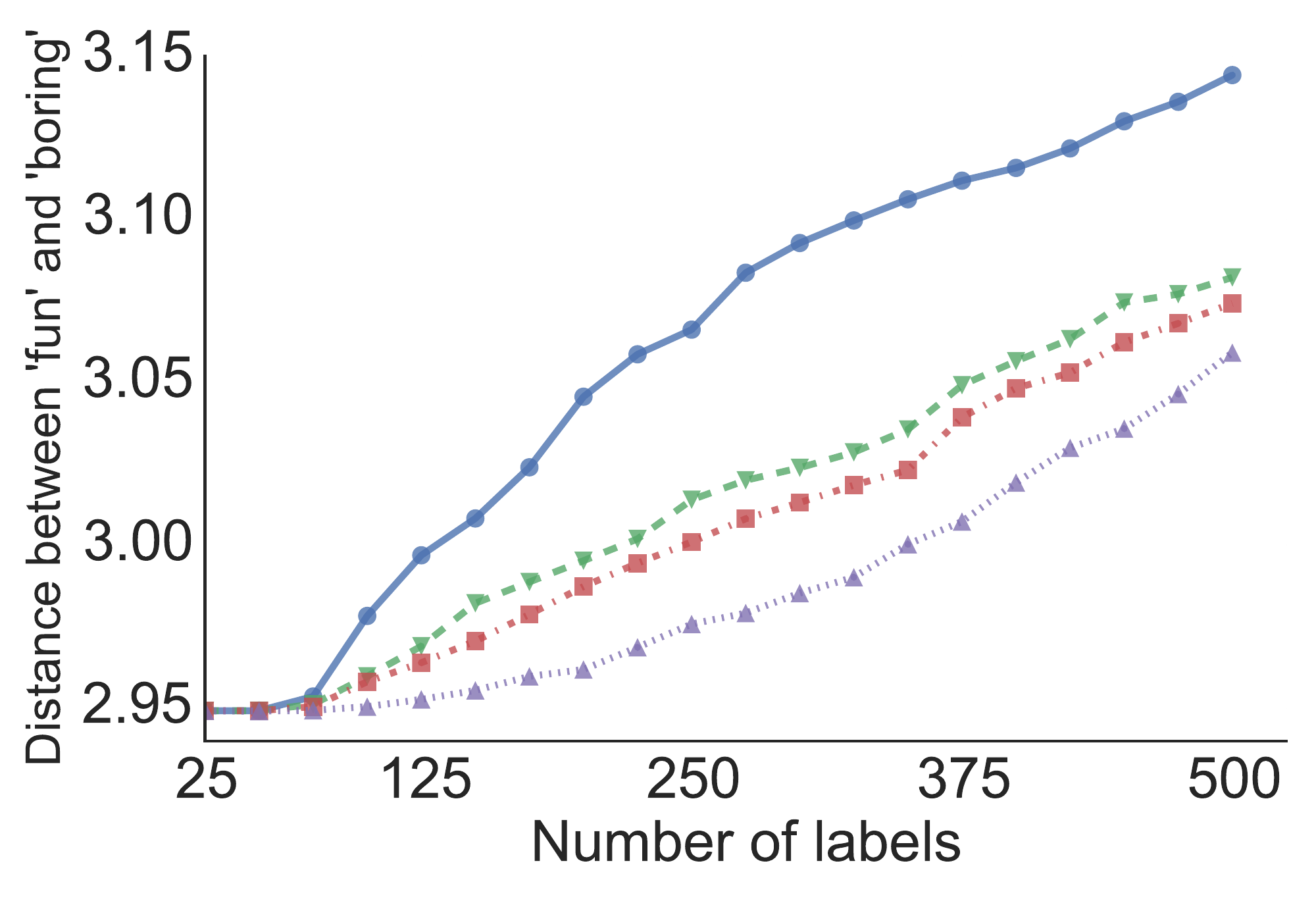}
    \includegraphics[width=\linewidth]{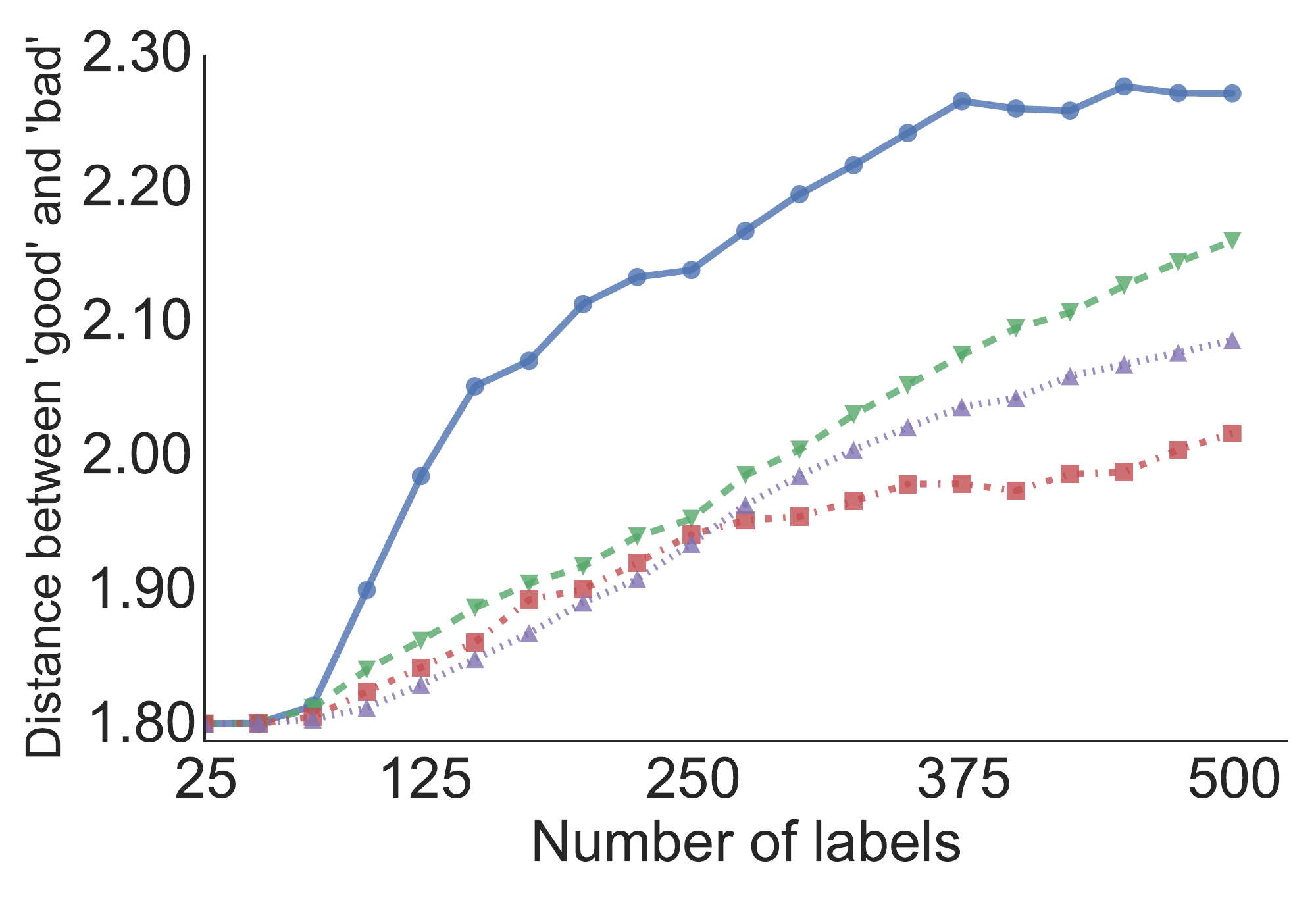}
    \includegraphics[width=\linewidth]{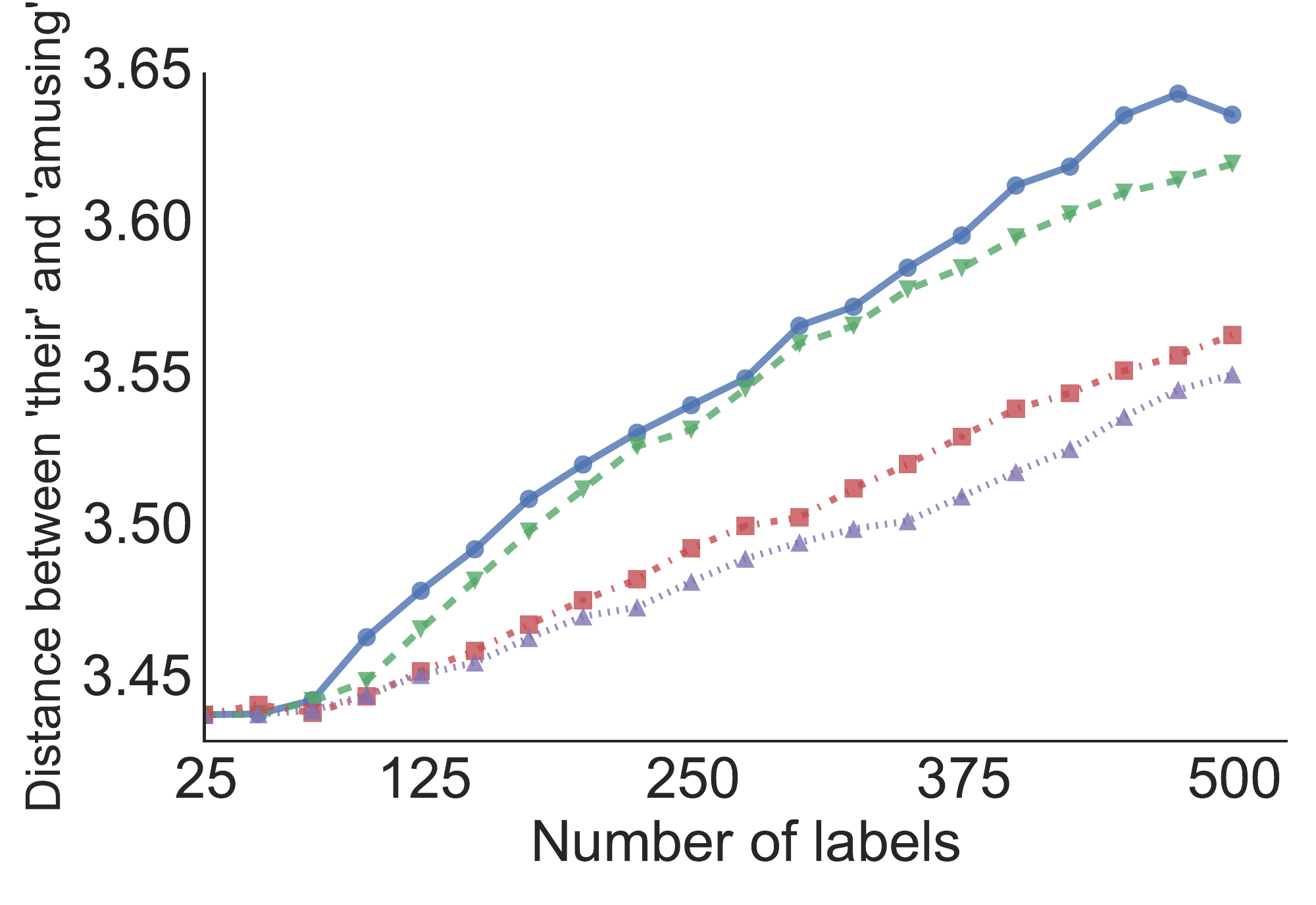}
\end{multicols}
%Ye_09: maybe drop 'EGL-sm' approach in the figure? 
%\vspace{-1.5em}
\caption{Results on the three sentence classification datasets. {\bf Top row:} number of labels versus accuracy. {\bf Bottom row:} number of labels versus the distance between tuned embeddings for selected pairs of informative words (with opposite polarity) for each dataset. The scale in this case, which captures the Euclidean distance in the embedding space, has only relative meaning.}
%Ye_12:I don't get "meaningful only in relative terms"
% bcw_13: I added this to address one of Matt's early comments. I just want to emphasize that the y-axis values here (distances in some crazy embedding space) don't really have any intrinsic or absolute meaning
\label{fig:embeddings-AL}
%\vspace{-9em}
\end{figure*}

\begin{figure*}[htbp]
%\vspace{-5em}
\begin{multicols}{3}
\includegraphics[width=\linewidth]{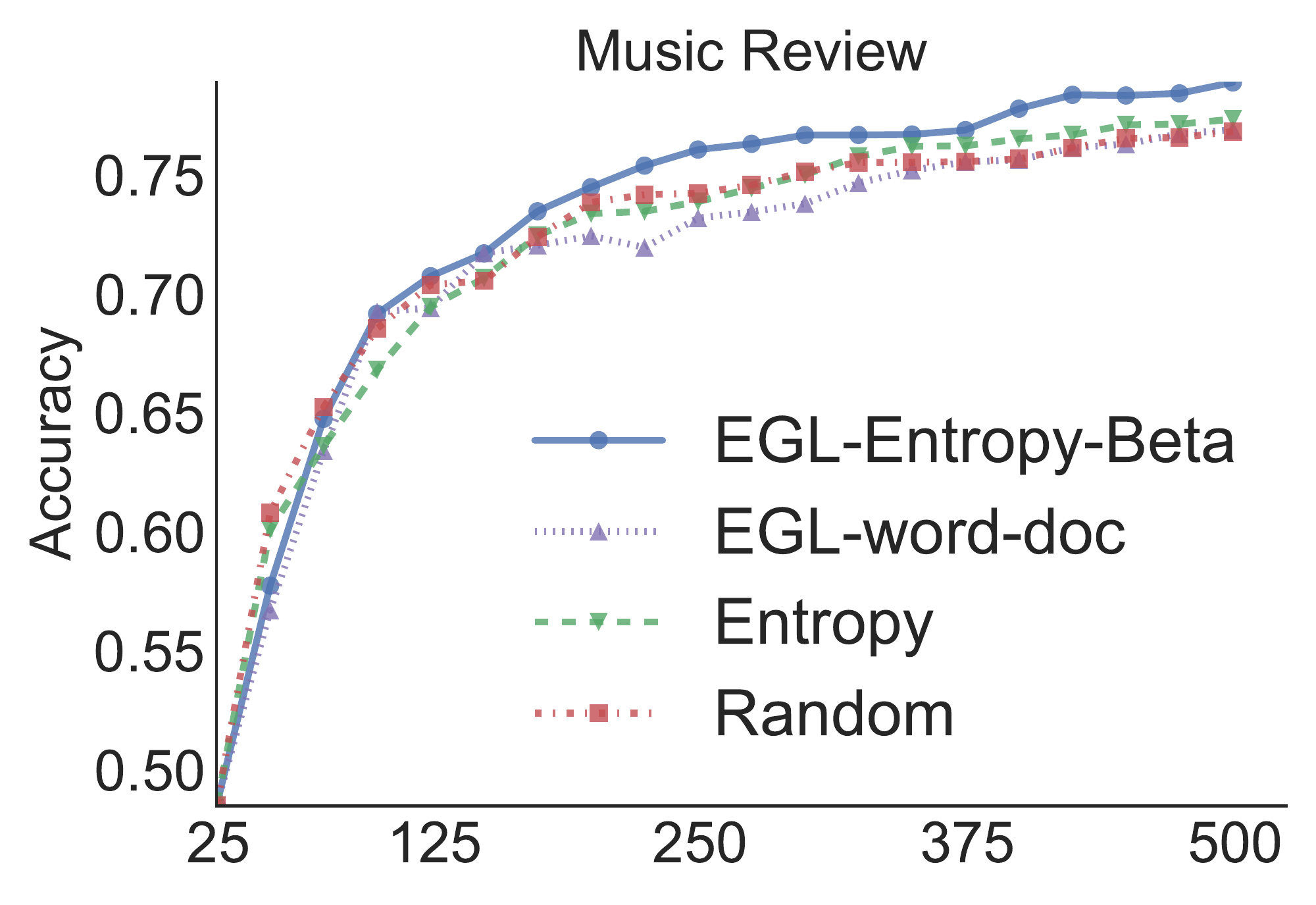}
\includegraphics[width=\linewidth]{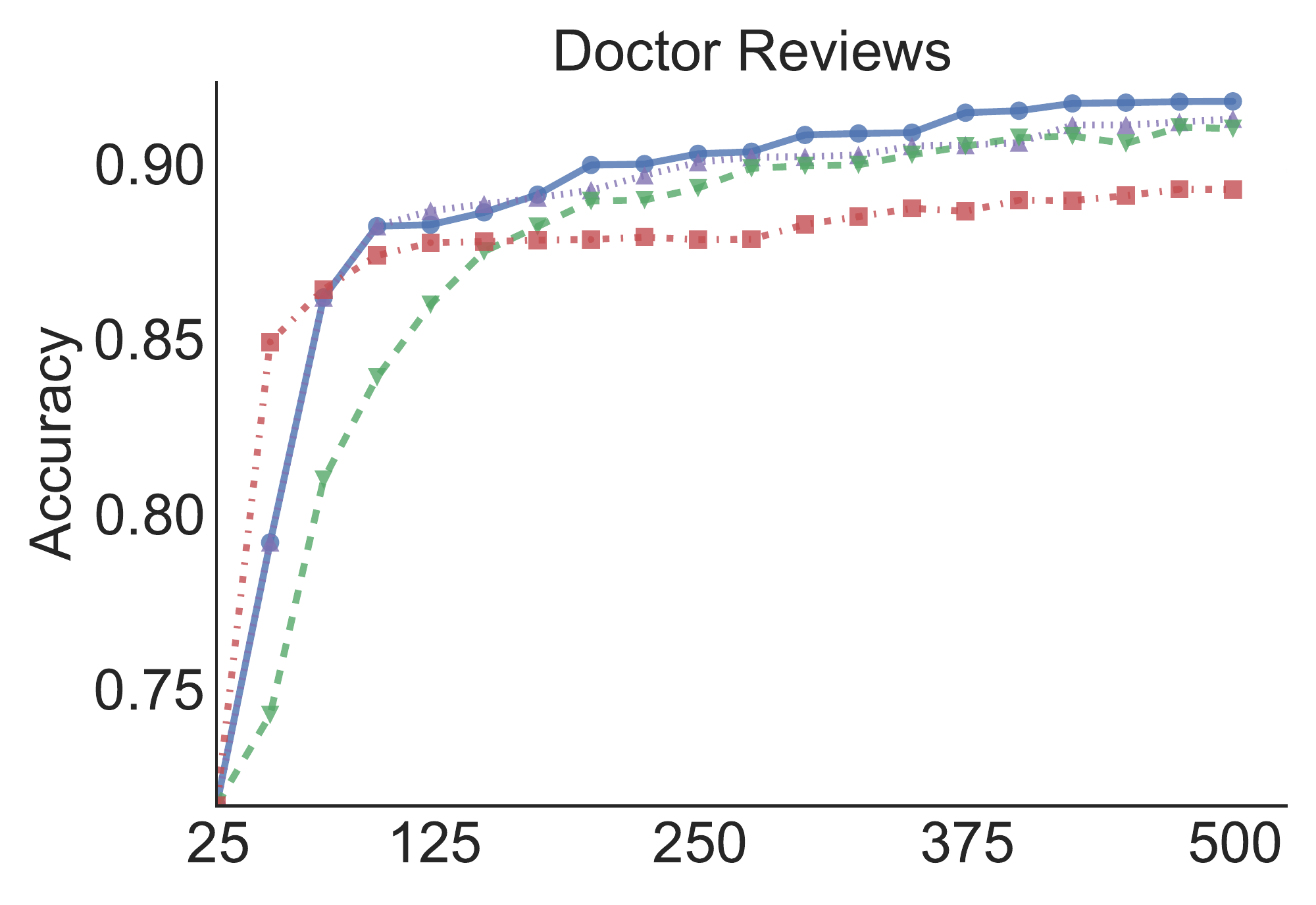} 
\includegraphics[width=\linewidth]{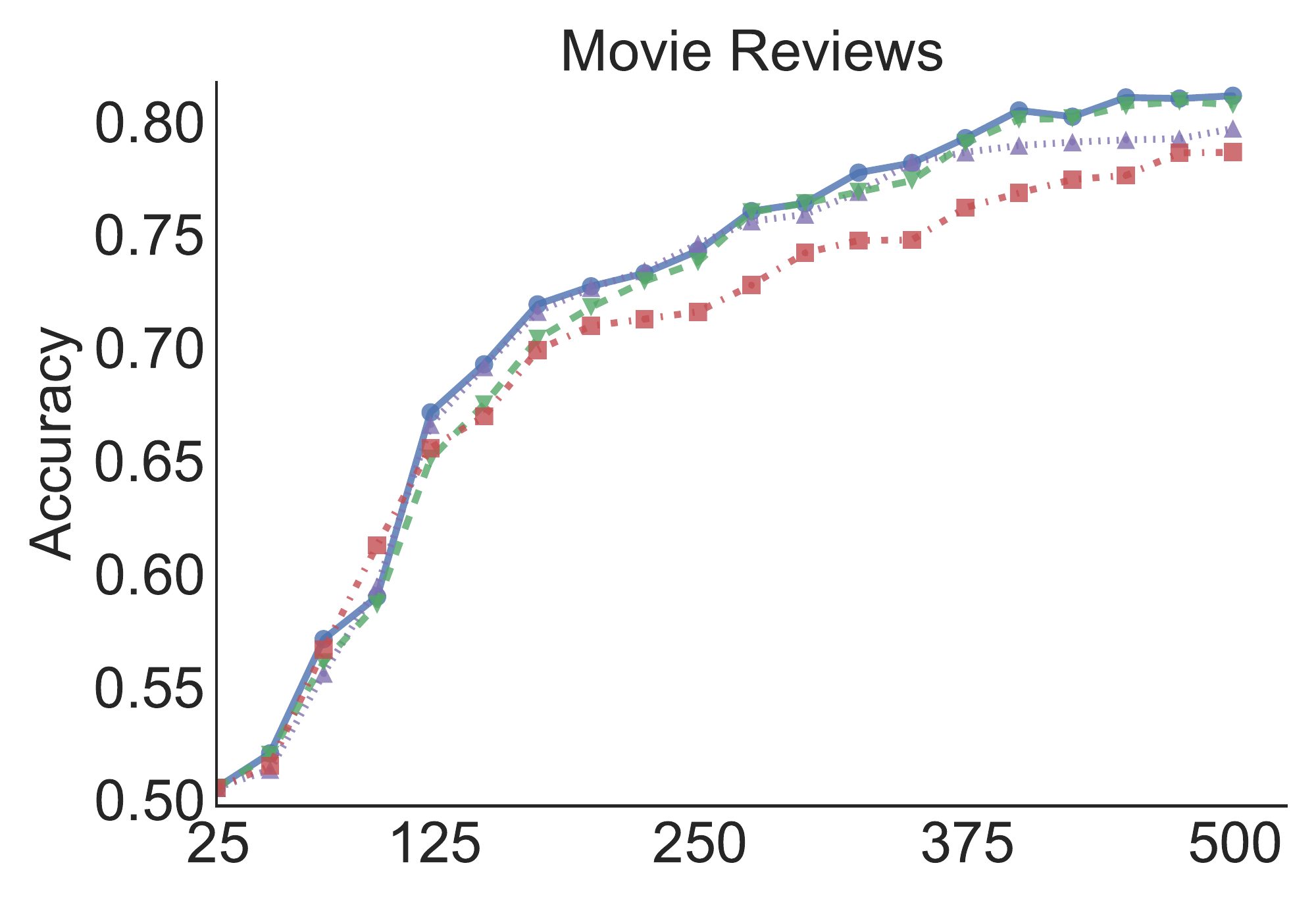}
\end{multicols}
%\vspace{3em}
\begin{multicols}{3}
    \includegraphics[width=\linewidth]{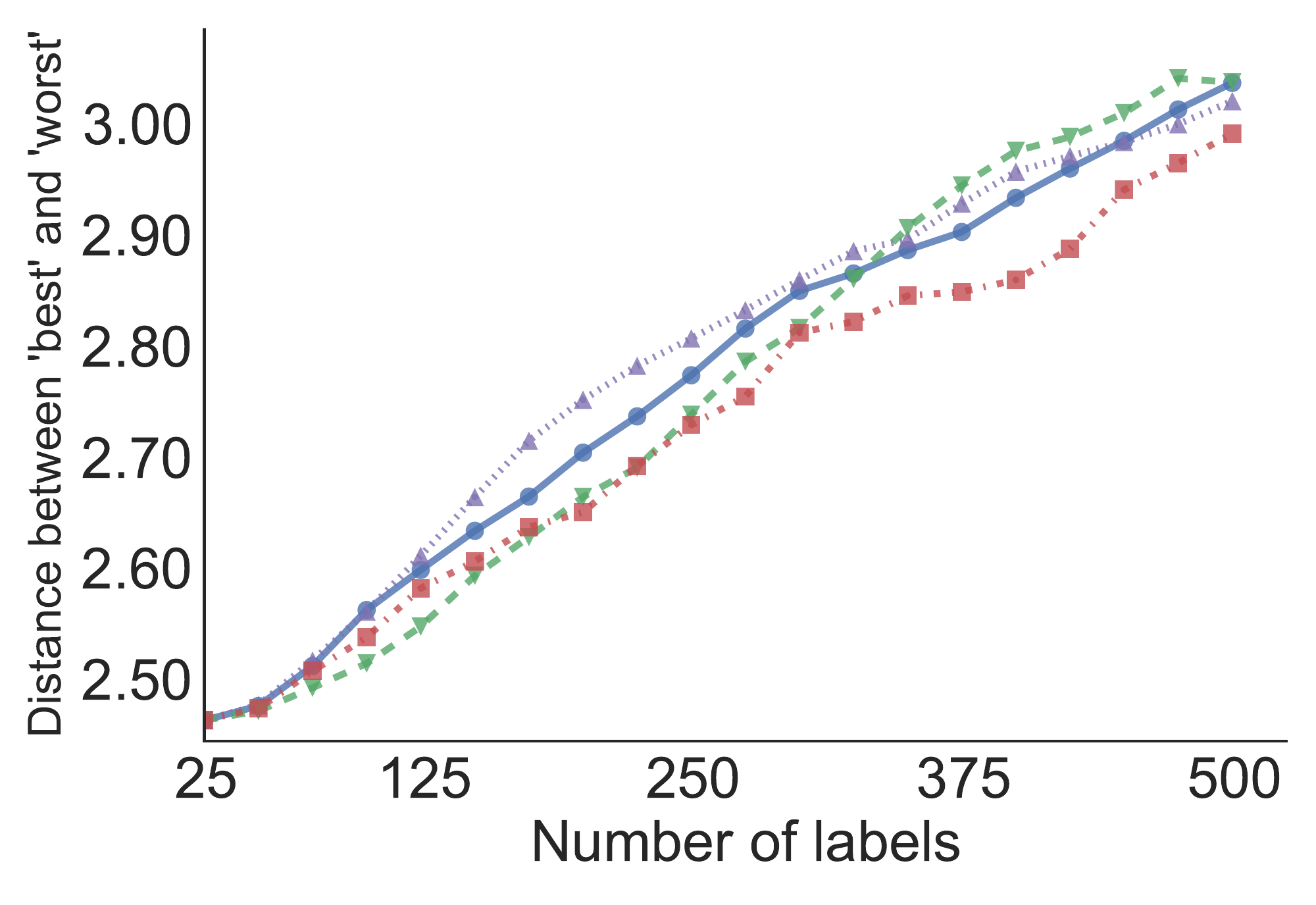}
    \includegraphics[width=\linewidth]{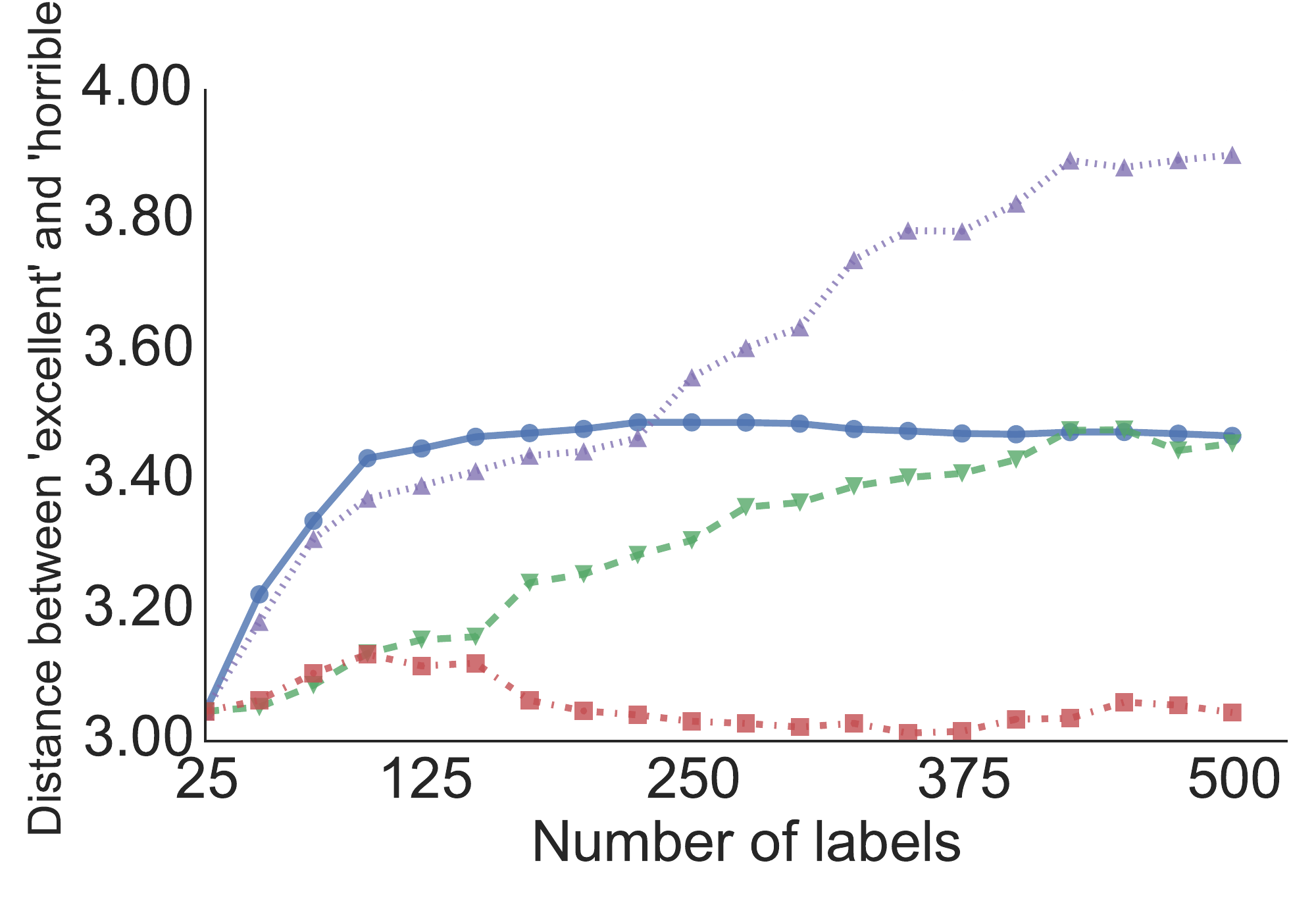}
    \includegraphics[width=\linewidth]{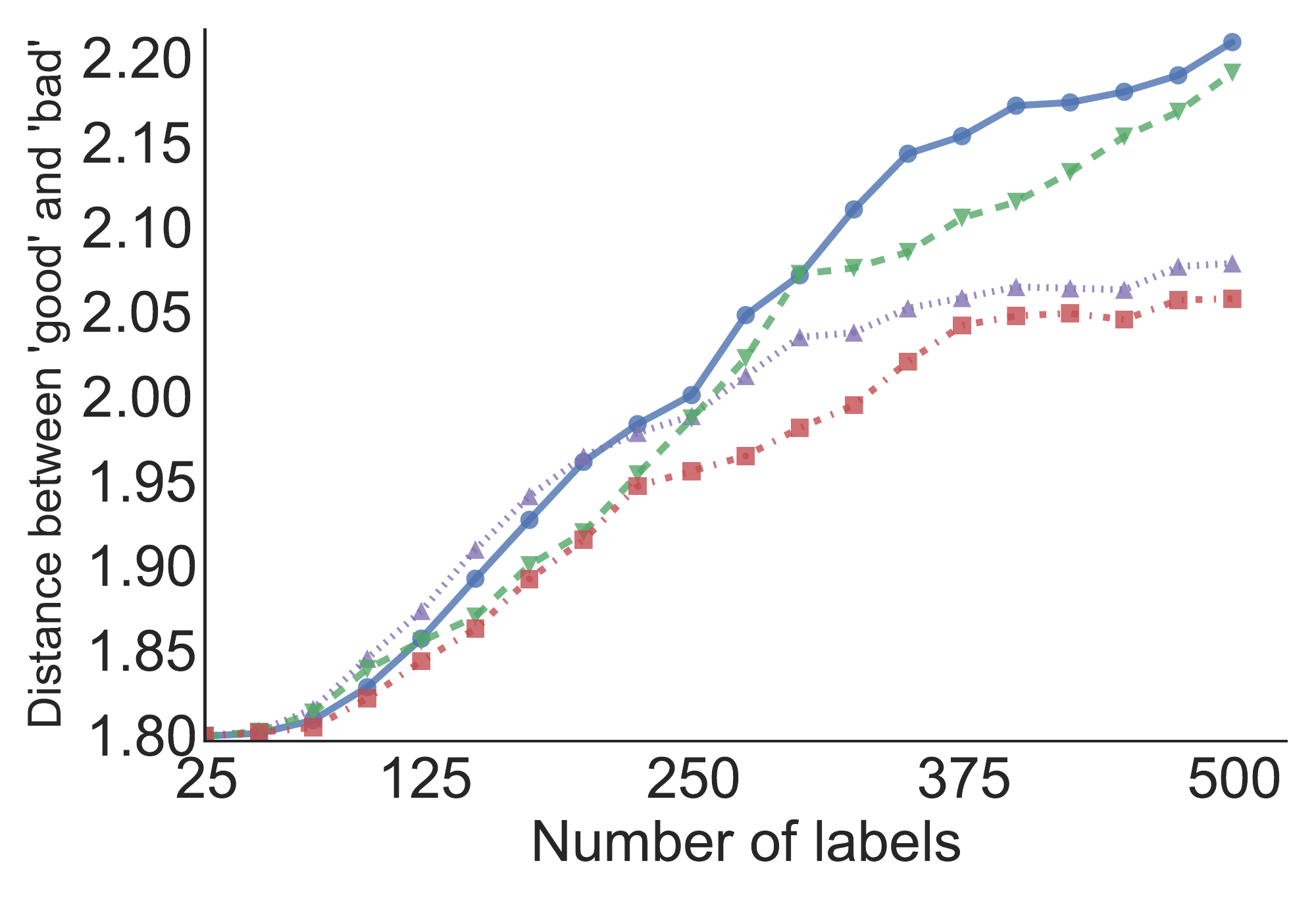}
\end{multicols}
%\vspace{-1.5em}
\caption{Results on the three document datasets. {\bf Top row}: number of labels versus accuracy. {\bf Bottom row}: number of labels versus the distance between tuned embeddings for selected pairs of informative words (with opposite polarity) in each task. \\
%\vspace{-5pt}
~\\
\noindent Methods that explicitly consider representation/embedding parameters more quickly push discriminative word vectors apart. Intuitively, the distances between the contrasting word-pairs increases quickly with both of the proposed EGL methods. However, recall that the {\em EGL-Entropy-Beta} method differs from {\em EGL-word-doc} in interpolating entropy along with expected updates to word gradients. As a result, we observe that {\em EGL-Entropy-Beta} method tends to shift from rising with {\em EGL-word-doc} at the start of learning, while later merging with the distances achieved by the {\em Entropy} method as learning progresses. This transition corresponds to first focusing on embeddings, and then later shifting emphasis to the entropy criterion.}
\label{fig:embeddings-AL-doc}
%\vspace{-2em}
\end{figure*}

We now report results. % for the sentence and document classification tasks, respectively. For the former task
For sentence classification, we use the simple variant of our method (\textit{EGL-word})
%Ye_11: should it be just EGL-word ??
which is more appropriate for short texts (since it is ultimately a max-operator over expected gradients for individual words). For  document classification, we also use the interpolated method, which considers expected gains both with respect to feature learning and in terms of instance-level uncertainty reduction. This method is more appropriate for longer texts.

%==========================================
\subsection{Sentence Classification Results}

{\bf Figure \ref{fig:embeddings-AL}} reports learning curves on the three sentence datasets. The proposed \textit{EGL-word} active learning method outperforms baseline approaches, performing especially well on sentiment analysis tasks (MR and CR). We believe this is due to our model rapidly learning more discriminative representations of words with opposing polarities.% (as shown in the bottom row of Figure \ref{fig:embeddings-AL}).

%This is attribute to that our EGL-word directly picks the instance that could greatly change the representation of discriminative words, which in turn could increase the performance more rapidly.  

\begin{comment}
\begin{figure}%[ht!]
\centering
\includegraphics[width=0.5\textwidth]{res_accuracy_CR.pdf}
\caption{Learning curve on CR dataset}
\label{res_CR}
\end{figure}
\end{comment}

To further illustrate this point, Figure \ref{fig:embeddings-AL}'s bottom row provides plots displaying the Euclidean distances between selected pairs of word embeddings induced using different AL strategies. In the customer review (CR) dataset, for example, we consider the embeddings of words `good' vs.\ `bad' and see that \textit{EGL-word} quickly pushes these embeddings apart. Similarly, on the movie review (MR) dataset, `fun' and `boring' are rapidly separated in embedding space. The subjectivity (Subj) detection task is less clear-cut. Here we picked words `amusing' and `their', because `amusing' strongly indicates subjectivity, while `their' is plainly neutral. As expected, \textit{EGL-word} quickly pushes these apart, though less rapidly than with the sentiment tasks. 

{\bf Table \ref{AUC}} reports Area Under Curve (AUC) scores for each learning curve from 25-500 labeled instances using trapezoidal rule~\cite{suli2003introduction}. We normalize AUC by the maximum possible for the range: $(500-25) * 1 = 475$.
\begin{table}%[h!]
\centering
%\footnotesize
\begin{tabular}{c | c c c c } 
%
% ML: one degree of precision sufficient; use same ordering of methods as labels in plot, and list accuracies all above distances, as in plot.
%
 \hline  \\[-.8em]
  & EGL-word & Entropy & Random & EGL-sm   \\ [0.25ex] 
 \hline
MR & \textbf{0.707}  & 0.690 & 0.681 &  0.667 \\ 
 \hline
 CR & \textbf{0.743} & 0.732 & 0.720 & 0.674 \\
 \hline
 Subj & \textbf{0.856} & 0.840 & 0.839 & 0.785 \\
 \hline
\end{tabular}
\caption{Area Under (learning) Curve (AUC) scores on sentence classification datasets; bold indicates  best results.}
\label{AUC}
\end{table}

%=================================================================
\subsection{Document classification results}

{\bf Figure \ref{fig:embeddings-AL-doc}} displays learning curves achieved on the document classification datasets, and {\bf Table \ref{AUC-doc}} reports the corresponding AUC scores achieved by each method on each dataset. Overall, the \textit{EGL-Entropy-Beta} outperforms other methods, demonstrating the value of explicitly selecting examples likely to improve representation level parameters. 

Results using the simple variant of \textit{EGL-word-doc} 
%\todo{Please be consistent throughout the paper on whether method names are italicized or not}
are mixed. 
In general it outperforms baselines only during the first several iterations of AL, but is later outperformed by entropy-based sampling. Our  intuition here is that narrowly focusing on improving feature representations provides early gains, but longer texts require attention to be shifted to instance-level uncertainty. And indeed, the proposed \textit{EGL-Entropy-Beta} method consistently performs more robustly, and tends to realize the best of both worlds, achieving rapid gains but also generally maintaining dominance over all AL iterations. 
%Ye_11
\begin{comment}
Though this is obvious on doctor review and music review dataset, we notice that the accuracy on movie review plot is not quite clear, since during many iterations, the several methods are very similar, so we give the numerical accuracy value for some iterations in {\bf Table \ref{Accuracy}} just for illumination. 
\end{comment}

\begin{comment}
\begin{table}[h!]
\centering
%\footnotesize
\begin{tabular}{c | c c c c c } 
%
% ML: one degree of precision sufficient; use same ordering of methods as labels in plot, and list accuracies all above distances, as in plot.
%
 \hline  \\[-.8em]
  & 25 & 125 & 250 & 375 & 500 \\ [0.25ex] 
 \hline
Random & 0.508 & 0.658 & 0.719 & 0.765 & 0.789\\ 
 \hline
Entropy & 0.508 & 0.654 & 0.741 & 0.793 & 0.810 \\
 \hline
EGL-word & 0.508 & 0.669 & \textbf{0.749}  & 0.789 & 0.800\\
 \hline
 E-E-B & 0.508 & \textbf{0.674} & 0.746 & \textbf{0.795} & \textbf{0.814}\\
 \hline
\end{tabular}
\caption{Accuracy at several number of labeled queries on MR document dataset. 
%Ye_13: Do we need to explain what "E-E-B" is? 
E-E-B refers to \textit{EGL-Entropy-Beta}.The best performance for each query number is in bold}
\label{Accuracy}
\end{table}
\end{comment}

\begin{table}%[h!]
\centering
%\footnotesize
\begin{tabular}{c | c c c c } 
%
% ML: one degree of precision sufficient; use same ordering of methods as labels in plot, and list accuracies all above distances, as in plot.
%
 \hline  \\[-.8em]
  & E-E-B & EGL-word-doc & Entropy & Random   \\ [0.25ex] 
 \hline
MR & \textbf{0.725} & 0.719 & 0.719 & 0.704 \\ 
 \hline
 DR & \textbf{0.893}& 0.889 & 0.877 & 0.878 \\
 \hline
 MuR & \textbf{0.736} & 0.718 & 0.725 & 0.726 \\
 \hline
\end{tabular}
\caption{Area Under (learning) Curves (AUC) scores on the three document datasets. E-E-B refers to \textit{EGL-Entropy-Beta}.}
\label{AUC-doc}
\end{table}
%by incorporating both EGL-word and entropy initially, and later focusing more on entropy, we could guarantee the robustness.

%In music review task, EGL-word is not even as good as random method. These observations verify our intuition that EGL-word alone might not suffice to represent the whole document. However, in \textit{EGL-Entropy-Beta}, by incorporating both EGL-word and entropy initially, and later focusing more on entropy, we could guarantee the robustness. 

Similar to {\bf Figure \ref{fig:embeddings-AL}}'s bottom row for sentence tasks, {\bf Figure \ref{fig:embeddings-AL-doc}}'s bottom row shows for document tasks how distances between selected word embeddings grow as more examples are collected. \textit{EGL-word-doc} and \textit{EGL-Entropy-Beta} consistently push the representations for the selected polar word-pairs apart more rapidly than other methods. However, recall that the {\em EGL-Entropy-Beta} method differs from {\em EGL-word-doc} in interpolating entropy along with expected updates to word gradients. As a result, we observe that {\em EGL-Entropy-Beta} method tends to shift from rising with {\em EGL-word-doc} at the start of learning, while later merging with the distances achieved by the {\em Entropy} method as learning progresses. \textit{EGL-Entropy-Beta} thus strikes a balance between this and refining the parameters at higher levels in the model, as evidenced by the superior classification performance seen in the top row of {\bf Figure \ref{fig:embeddings-AL-doc}}. Maintaining a narrow focus on embeddings only ultimately results in comparatively poor performance in the case of document classification.

\section{Conclusions}

The importance of \emph{representation learning} \cite{bengio2009learning} with neural models motivates exploring new, representation-based active learning (AL) approaches with neural models. To this end, we proposed a new AL strategy for CNNs that is specifically designed to quickly induce discriminative, task-specific representations (word embeddings), thus improving classification. We showed that this approach outperforms baseline AL strategies across sentence and document classification datasets considered, and that such discriminative word embeddings can be rapidly induced.

We believe that these encouraging results will help to stimulate further research on active learning tailored to deep/hierarchical architectures. Our own future work will include generalize the similar AL strategies to other neural models such as recurrent neural network %Ye_1117
and improving the modeling strategy for $\gamma_t$ (the parameter governing relative emphasis on representation vs. instance-level uncertainty), perhaps based on reinforcement learning. We also envision augmenting the model to optimize instance selection in terms of refining additional intermediate layer representations in deeper networks.

\section{Acknowledgments}
 This research was supported in part by IMLS grant RE-04-13-0042-13 and the Foundation for Science
and Technology, Portugal (FCT), through contract UTAPEXPL/EEIESS/0031/2014. Any opinions, findings, and conclusions or recommendations expressed by the authors do not express the views of any of the supporting funding agencies.
%specifically designed for sentiment analysis. 

%\bibliography{acl2016}
\bibliography{reference.bib}
\bibliographystyle{aaai.bst}
\end{document}